%% file: main.tex
\documentclass[10pt,twocolumn,letterpaper]{article}

\usepackage[pagenumbers]{cvpr} 

\usepackage{graphicx}
\usepackage{amsmath}
\usepackage{amssymb}
\usepackage{booktabs}

\usepackage[pagebackref,breaklinks,colorlinks]{hyperref}

\usepackage[capitalize]{cleveref}
\crefname{section}{Sec.}{Secs.}
\Crefname{section}{Section}{Sections}
\Crefname{table}{Table}{Tables}
\crefname{table}{Tab.}{Tabs.}

\def\cvprPaperID{*****} 
\def\confName{CVPR}
\def\confYear{2022}

\begin{document}

\title{Analyzing Compression Techniques for Computer Vision}

\author{Imran Khan\\
PhD ORIE\\
{\tt\small iakhan@utexas.edu}
\and
Maniratnam Mandal\\
PhD ECE\\
{\tt\small mmandal@utexas.edu}
}
\maketitle

\begin{abstract}
   Compressing deep networks is highly desirable for practical use-cases in computer vision applications. Several techniques have been explored in the literature, and research has been done in finding efficient strategies for combining them. For this project, we aimed to explore three different basic compression techniques - knowledge distillation, pruning, and quantization for small-scale recognition tasks. Along with the basic methods, we also test the efficacy of combining them in a sequential manner. We analyze them using MNIST and CIFAR-10 datasets and present the results along with few observations inferred from them.  
\end{abstract}

\section{Introduction}
\label{sec:intro}
Neural network compression is an important issue for improving the usability of networks in smaller and less powerful devices. As the size of deep networks increases, they become increasingly more demanding in their processing power and storage requirements. Some of the largest neural networks built for vision tasks contain parameters on the scale of billions. As the growing trend is that neural networks are increasingly used in mobile phones, it is sometimes infeasible for extremely large networks to be trained, or even deployed. 

In addition to storage and memory concerns with over-parameterized neural networks, there are other concerns that would warrant smaller, more usable networks. It is important to note the speed and latency concerns of networks are just as important. Additionally, it is estimated that the GBT-3 network, with 175 Billion parameters, emitted a total of 552.1 tons of carbon dioxide in training \cite{carbonemissions}. Thus, if reducing the size of networks can lead to increased energy efficiency, there are potential environmental and ethical reasons for wanting to reduce the size of extremely large networks.  

One simple solution to combat this problem is to make smaller networks with much fewer and manageable parameters. While this solution would decrease the size of the networks, this comes at the cost of decreased model performance. This would not be favorable for many recognition based applications, be it auto-navigation, image captioning, visual question answering, or face recognition-based security measures. Similarly, another simple solution would be to “outsource” the model to a server. This would mean processing the model on a server, which would require sending large amounts of data back and forth between the mobile phone and the server. This is also not a perfect solution to our problem, because this would require a great internet connection for the data to be sent, and would incur unnecessary delays that slows down the processing. For example, imagine having to wait every time you wanted to use facial recognition to unlock your phone. 

These solutions, although feasible, are not sufficient. The field of network compression has introduced several techniques for mitigating this. The goal of neural network compression is to obtain the same or similar generalization performance by efficient storage, optimization, computation, and often reduction in network size. The reduction in storage and computation leads to critical improvements in real-time applications such as in mobile devices, smart wearables, online learning, etc. Some of the recent major techniques in network compression are – parameter pruning and quantization, low-rank factorization, transferred/compact convolutional filters, and knowledge distillation \cite{netComp2}. 

According to \cite{lecture8}, these methods can be categorized into two common streams: the "transfer" stream and the "compress" stream. Techniques in the transfer stream is focused on training a new small network, whereas techniques in the compress stream are focused on decreasing the size of the model during or after training. As the techniques have been developed independently, they can be used in conjunction and to complement each other. The goal of this project is to analyze and combine these techniques to achieve higher levels of compression while preserving performance. The major gap that this project is hoping to fill is the gap between the two streams. We would like to show empirical results of neural network compression techniques that can be done through a combination of transferring knowledge to a smaller network and simultaneously compressing the networks. 

Our paper is organized as follows: the following section contains a review of the literature that has already been done on the subject of neural network compression; Section 3 discusses the methods we choose to use in our experiments; Section 4 goes into more detail about the details of our experimental setup; Section 5 talks about our results; and we end with a conclusion in Section 6.

\section{Related Work }
\label{sec:Related Work}

\subsection{Quantization}
Quantization has been widely used as a compression technique across all fields of engineering sciences. Reducing the number of bits to represent each weight can lead to reduced memory and increased computation \cite{quant1}. \cite{quant1_5} aimed to decrease the size of the network through storing with a smaller format. In this paper, the authors test three benchmark data sets with three different storing formats: floating point, fixed point, and dynamic fixed point. They find that very low precision storage is good enough for running trained networks, and that it is also sufficient for training them. 

Binary weight neural networks have also been explored \cite{quant3}, although the performance of such networks take a major hit in the case of deep CNNs. \cite{quantizingdeep} introduces Quantization Aware Training, which is a method of helping the accuracy of a model be less affected with Post Training Quantization, by accounting for the quantization loss in the training phase of the model. This will be discussed in more detail in Section 3.3. 

\subsection{Pruning}
Early pruning based methods reduced the number of connections among the layers based on the loss function \cite{netComp2}. Pruning methods also include removing redundant neurons, remove redundant connections, quantizing and encoding weights, and parameter sharing. HashedNets uses a low-cost hash function to store and group weights for parameter sharing \cite{hashnet}. Some compact DNNs are trained with sparsity constraints during optimization for regularization. Pruning with regularization can require a large time to converge and usually involve intricate parameter tuning, thus increasing the training and inference time. Optimal Brain Damage \cite{optimalbraindamage} also shows a more robust method of pruning. By removing unimportant weights from a network, they claim they can get improved generalization, require fewer training  data points, and improve the speed of learning. This technique takes second-derivative information to balance between network complexity and training set error. 

\subsection{Knowledge Distillation}
Knowledge Distillation (KD) was originally introduced as a training procedure that aims to transfer the knowledge learned by a deeper network (Teacher) to another relatively shallower one (Student), while maintaining the same level of performance \cite{hinton2015distilling}. The motivation behind this technique is to reduce the computational complexity of some operations, or compression of large networks, such that devices with limited capacity (e.g. smartphones) could benefit from the achievements of deep learning. In KD, the student network is trained not with the original hard binary labels of the dataset, but instead with the soft labels taken from the output probability of the teacher network. The student model learns to reproduce the output of the teacher model which is used for training. In \cite{hinton2015distilling}, an ensemble of teacher networks was compressed into a shallow student using the softmax output of the teachers. FitNets \cite{fitnet} compress thin deep networks into wide shallow ones while preserving the performance, and have also been used to learn full feature maps from the teacher. There have been extensions of KD where students have been trained to approximate Monte Carlo teacher \cite{kd2}, trained to represent knowledge in higher neural layers \cite{kd3}, or using multiple networks of decreasing depth in between \cite{takd} to transfer knowledge from teacher to assistant to student. Although KD can compress very deep models into much shallower ones, it takes the largest hit in performance among all the compression techniques due to its limited representation capabilities. 

\subsection{Combined Methods}
In \cite{quant2}, the authors use a combined pipeline of Pruning, Quantization, and Huffman Coding. They first prune the model, then use weight sharing and post-training quantization to further reduce the size of their model by 39-50x compression. They manage to achieve this compression with no loss in the accuracy of the model. 

\subsection{Other Methods}
 In \cite{predictingparameters}, the authors demonstrate that there is significant redundancy in the weights of large deep learning networks. With a few weight values for each feature they claim it is possible to accurately predict the few unknown weights. Additionally, they show that not only can the parameter values be predicted, but many of them are simply not necessary. In this paper, they train several different networks by learning only a fraction of the parameters and predicting the remainder. In the best cases they are able to predict more than 95\% of the weights of a network without a decrease in accuracy.

In \cite{reallydeep}, the authors show that shallow fully connected models can learn the same functions, regardless of how complex, that were learned by deeper models and attain the same performance of the deep models. In some cases the shallow neural networks learned these functions using the same number of total parameters as the deeper model. They evaluate the method on the TIMIT phoneme recognition task and successfully train shallow feed forward networks that perform just as well as deep convolutional networks. They suggest that there exist better algorithms for training shallow feedforward nets than those currently available.

Other methods of network compression include using low-rank filters to accelerate convolution \cite{other1}, using transferred convolutional filters \cite{other2}, attention-based mechanisms to reduce computations \cite{other3}, randomly dropping and bypassing layers with identity function \cite{other4}, etc.

\section{Methods}

In this section, we will introduce the methods we use for our experiments. We introduce the baseline model, the compression techniques we use, and the motivation for these selections.

\subsection{Baseline Network Architecture}

\begin{figure}[t]
\begin{center}
\includegraphics[width=0.7\linewidth]{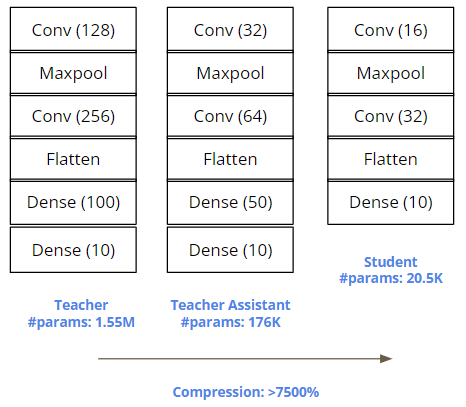}
\caption{\scriptsize{\textbf{Model Architectures for MNIST} }}
\vspace{-2em}
\label{fig:mnistmodel}
\end{center}
\end{figure}

\begin{figure}[t]
\begin{center}
\includegraphics[width=0.7\linewidth]{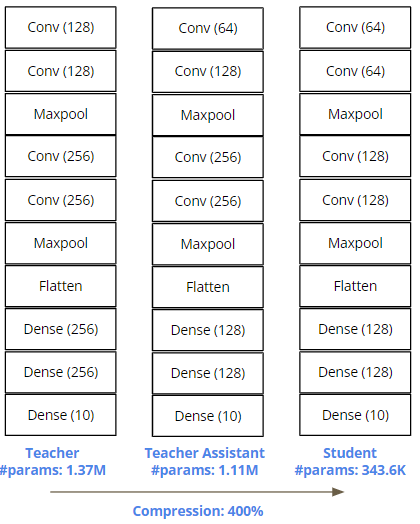}
\caption{\scriptsize{\textbf{Model Architectures for CIFAR10} }}
\vspace{-2em}
\label{fig:cifar10model}
\end{center}
\end{figure}

The proposed network architectures are shown in Figures \ref{fig:mnistmodel} and \ref{fig:cifar10model}. 

We use two different base models for MNIST \cite{mnist} and CIFAR-10 \cite{cifar10} datsets. Because we wanted to test the effect of network compression on convolutional layers and dense layers separately, both the models were created accordingly. The base MNIST model has two convolutional layers of size 128 and 256, followed by a hidden dense layer of size 100. Whereas, the CIFAR-10 base model has 4 convolutional layers of sizes 128,128,256, and 256, followed by two hidden layers of size 256. The MNIST and CFIAR-10 base models have aorund 1.55M and 1.37M trainable parameters respectively.

\subsection{Knowledge Distillation}\label{sec:KDistill}
To test the efficacy of knowledge distillation, we tray both single-step and multi-step distillation methods. In the former case, we train a student model with the output generated from the teacher (baseline) model as labels, while in the latter case, we introduce a teacher assistant model \cite{tadistill} as well. For both the techniques, the student (compressed) model was kept the same. The MNIST student has two convolutional layers of size 16 and 32, followed by flattening and the output dense layer. There was no hidden dense layer in this case. Same as the corresponding teacher, the CIFAR-10 student has 4 convolutional layers (size 64, 64, 128, and 128) and two dense hidden layers of size 128. The MNIST and CIFAR-10 students have around 20.5K and 343.6K parameters, leading to $>7500$\% and $>400$\% compression respectively, when compared to their corresponding teacher models. 

For testing the efficacy of multi-step distillation, we train a Teacher Assistant model with labels generated by the teacher, and then train the student using the teacher assistant labels. The Teacher Assistant model is smaller than the Teacher but larger than the Student model for both cases. The MNIST TA has around 176K parameters, and the CIFAR-10 Teacher Assistant has around 1.11M parameters.
For both single and multi-step distillation, we also study the effect of distillation temperature on student accuracy.

\subsection{Quantization}

We examined two methods of quantization of our networks: Post-Training Quantization and Quantization Aware Training. The first quantization technique we explored was Post Training Quantization. Post Training Quantization is performed on a pre-trained model, and it simply reduces the bit-precision of the weights and activation functions within the network. For example, a network which is normally stored with 32-bit float weights can be instead stored with 8-bit weights, for a theoretical four times decrease in storage size of the network \cite{quantizingdeep}. 

The second quantization technique we used was Quantization Aware Training is a method of training a neural network that is meant to help the model accuracy downstream when later performing Post-Training Quantization. Quantization Aware Training works by emulating lower precision steps in the forward pass, while keeping the backward pass the same. This way, the optimizer can account for the quantization error in the training phase of the network, and can modify weights step-wise accordingly. Then, at the end of training the network, we can store the weights with Post-Training Quantization and should theoretically, get better validation scores than we would have without the Quantization Aware Training \cite{quantizationawaretraining}.

\subsection{Pruning}
We compared two different pruning strategies on both of our base models - global and local pruning. For each, we explored both constant sparsity and polynomial decay with changing sparsity training schedules. All the pruning strategies we used in our project was unstructured magnitude-based, which involved iteratively pruning low magnitude, thus less important, weights during training until the desired sparsity is reached.

Polynomial decay involves two different sparsity parameters - the initial sparsity and the final desired sparsity. We trained the models by varying them in three different ways - 0 to 80\%, 0 to 50\%, and 50 to 80\% weight sparsity. In constant decay, the sparsity remains the same as the model is fine tuned. In this case, we compare three different pruning outcomes with 20\%, 50\%, and 80\% sparsities. 

Along with global pruning, we also experimented with partial local pruning. We wanted to study the effects of pruning on CNN and dense parts of a network. Hence, for each dataset baseline, we prune the convolutional layers and dense layers separately with both polynomial decay and constant pruning strategies as described before. For all the pruning methods, the pruned trained network was fine tuned for two epochs. 

\subsection{Combined Techniques}
We also explored the combination of different compression techniques to study their efficacy. Primarily, we sequentially combine knowledge distillation with pruning and quantization separately. For the first set of experiments, we first compress the teacher using knowledge distillation and then apply pruning and quantization separately. And for the second set of experiments, we first prune and quantize the teacher, followed by training the student. Although the second experiment does not compress the (final) student model and thus not contribute to further compression, we were curious to study the effects of teacher pruning on distillation.

\section{Experimental Setup}

In this section, we explain the details of how we carry out the experiments, including the implementation and the experimental design. 

\subsection{Datasets}

The datasets we used for experimentation on the model were the MNIST Digits dataset and the CIFAR-10 image recognition dataset. The MNIST dataset consists of 60000 training images of labeled handwritten digits, and 10000 testing images. The digits are centered and size normalized, and each image is represented by a 28x28 grid of pixels. There are a total of 10 labels, one for each digit 0-9 \cite{mnist}. 

The CIFAR-10 dataset consists of 60000 training images and 10000 test images. This dataset also has ten labels: airplanes, automobiles, birds, cats, deer, dogs, frogs, horses, ships, and trucks. The images here are larger than the MNIST images. They are represented as color images in a 32x32x3 tensor \cite{cifar10}. We included this dataset as part of our experiment because we were interested in how the additional dimension of color might affect the networks compression ratios and accuracies. 

\subsection{Implementation}

All experiments were done in Python using Tensorflow libraries and a Keras framework. We used the tensorflow\_model\_optimization package to help with quantization aware training. As there were no readily available packages for knowledge distillation in TensorFlow, we had to create our own class for distillation. We used Google Colab for our programming environment.  

For the implementation of our algorithms, we used five epochs for our main models and teacher models, and when we trained student models for knowledge distillation, we used three epochs. We used an ADAM optimizer for all models and it's default learning rate of 0.001. We used the Sparse Categorical Cross Entropy Loss as our loss function. For reproducability, we set a random seed of 1234. Before training, we also normalized all pixel values which originally given to a value between 0 and 1. 

\subsection{Evaluation Metrics}

To evaluate our experiments, we looked at three main metrics. The first metric, was the testing accuracy of each model. With each of the datasets, we held out the test set as to not contaminate the model, and for training we incorporated a 0.1 validation split. The testing accuracy  can be measured as a percent of the test set correctly predicted.

The second metric we looked at was the storage size of the model. To find the storage size, we saved all models in a TFLite format and read the total number of bytes that were saved in the saved model's file path. 

The final metric we observed was called the efficacy metric. This was simply calculated as a ratio of the model accuracy to the size of the model. While this last metric is not very intuitive, we wanted to include it as a way of evaluating a compression technique with a mix of how its model's accuracy changes with the changes in size.

\section{Results}

In this section, we will go over the results of the different experiments we ran. 

\subsection{Knowledge Distillation}
\begin{figure}[t]
\begin{center}
\includegraphics[width=1\linewidth]{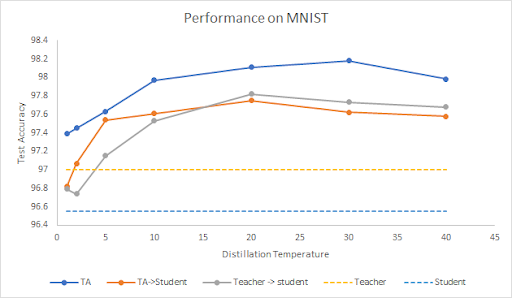}
\caption{\scriptsize{\textbf{Evaluating KD on MNIST: } The performance of the Teacher, TA, and Student models on the MNIST data set \cite{mnist} for different distillation temperatures.}}
\label{fig:mnistKD}
\end{center}
\end{figure}

\begin{figure}[t]
\begin{center}
\includegraphics[width=1\linewidth]{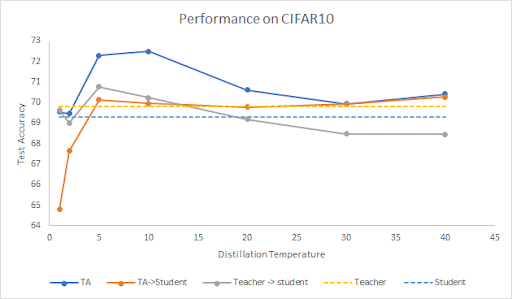}
\caption{\scriptsize{\textbf{Evaluating KD on CIFAR-10: } The performance of the Teacher, TA, and Student models on the CIFAR-10 data set \cite{cifar10} for different distillation temperatures.}}
\label{fig:cifarKD}
\end{center}
\end{figure}

Figures \ref{fig:mnistKD} shows the performance of the teacher, teacher assistant, and student models (Sec. \ref{sec:KDistill}) on the MNIST \cite{mnist} data set for distillation temperatures $T = \{1,2,5,10,20,30,40\}$. We notice that distillation almost always leads to an improvement for both the single step and multi-step cases. When compared between the two, we observe that introducing TA improves the performance of the student for lower distillation temperatures, whereas for high temperatures, the performance remains almost the same. Furthermore, the performance of TA is slightly better than the student models, though the student model contains only about $11$\% number of parameters compared to TA. This points to the high compression capabilities of knowledge distillation. The optimal distillation temperature seems to be around 15--30.

Similar experiments on the CIFAR-10 models (Fig. \ref{fig:cifarKD}) shows a slightly different trend. Here too, for most cases, the distilled models perform better than the teacher. But, unlike the MNIST case, here the TA leads to better performance for higher distillation temperatures, and for lower temperatures, TA does no help. For both the plots, we notice the performance to be higher in the middle and lower for low and high temperatures. Low \textit{T} results in harder labels and hence lower accuracy, whereas high \textit{T} leads to softer labels but with larger entropy, which may have caused the lower accuracy for high temperatures.

\subsection{Pruning}
\noindent\textbf{Global Pruning: } Table \ref{tab:glob_pruning} shows the performance of the unstructured pruning for both the polynomial decay and constant sparsity schedules. The percentage improvements over the original baselines have been presented. We observe that performance increases with increase in sparsity. For these set of experiments, the fine tuned pruned models always performed better than the original baselines. Similar to quantization, we noticed a huge improvement in model size reduction, although the accuracy might have slightly changed.
\input{latex/tables/global_prune}

\noindent\textbf{Local Partial Pruning: } 
We experimented with pruning the dense layers and convolutional layers separately. The results of local partial pruning on the CIFAR-10 baseline is presented in Table \ref{tab:local_pruning}. We notice a slightly better performance when pruning the convolutional layers, which might be due to the innate higher sparsity in them compared to the fully connected layers. Also, comparing to Table \ref{tab:glob_pruning}, we observe that, when compared to global pruning, local pruning is generally less advantageous in terms of compression efficacy (performance/size).   
\input{latex/tables/local_prune}

\subsection{Quantization}
\input{latex/tables/quantization}

The results of the quantization experiments are shown in Table \ref{tab:quant_results}. What we can look at first is the accuracies and sizes of our baseline models- in the MNIST case, the baseline model performs well with an accuracy of 97.74, with a size of 5.923 MB. In the CIFAR-10 case, the baseline model has an accuracy of 70.42 and a storage size of 5.22 MB. 

The first observation we can take notice of is the results of the Quantization Aware Training experiment. The accuracy in the case of MNIST actually improved from the original model, while the accuracy of CIFAR-10 went down, but not significantly. Overall, the accuracies here did not change much, but the size of the model decreased significantly. When we applied the Quantization Aware Training, we also apply a 8-bit Post Training Quantization. The size of the model thus decreases to 1.494 MB in the case of MNIST, and it decreases to 1.332 MB in the case of CIFAR-10. These decreases represent an approximate compression ratio of 4x. 

The next thing we can notice from the table are the the changes in the original model when just Post Training Quantization is applied. We can notice that the accuracies of the quantized models in MNIST remain almost exactly the same as the original model (97.75 and 97.74 in the 8-bit and the 16-bit models, respectively, vs 97.74 in the original model), and that the accuracies of the CIFAR-10 16-bit quantized model is exactly equal to the accuracy of the CIFAR-10 original model. We can, however, see a very slight decrease in accuracy of the the CIFAR-10 8-bit model, but it is not very significant. We can also observe the storage size differences in all of these models. As expected, the 8-bit models empirically show an approximate compression ratio of 4x from the original 32-bit model, and the 16-bit models empirically show an approximate compression ratio of 2x. 

Another interesting observation is the comparison between the Quantization Aware Trained network and the Post-Trained models. It is interesting to note that the Quantization Aware Trained model has a higher accuracy in the MNIST case, which is what is expected from having the Quantization Aware forward passes incorporated in the optimizer. However, we don't see that same pattern in the CIFAR-10 case, where the Quantization Aware accuracy is actually less than both of the Post-Trained models' accuracies. The model sizes remain almost exactly the same, which is expected when they both have the same total number of parameters and the same bit-precision storage. 

\subsection{Combined Techniques}
\noindent\textbf{KD + Pruning: } Table \ref{tab:kd_pruning} shows the comparison between the individual techniques with the combined one. We notice that, although the accuracy of the prune student decreased after pruning, it was accompanied with a huge reduction in size. So, for simple models and data sets, combining the two methods may lead to huge improvements. 
\input{latex/tables/kd_prune}

\noindent\textbf{KD + Quantization: } From Table \ref{tab:quant_results}, we observe that the performance of the Distilled Model (Teacher to Student) is exactly equal to the Distilled Model (Quantization Aware Trained Teacher to Student). Essentially, this shows that there was no difference in that distillation process when we gave it the original teacher model or the quantization aware teacher model. What we can see, however, is the large difference in the sizes of the student model when compared to the teacher model, and the very small difference in the accuracy. In the MNIST case, the storage size of the model goes from 5.923 MB to 0.081 MB, which is a compression ratio of almost 75x. This shows that a much smaller network can be learned from a significantly larger network.

Furthermore, we can sequentially add on post training quantization to the distillation process. This even further decreases the size of the model by another 4x, without affecting the accuracy. In the MNIST case, the final compression ratio after going through both distillation and 8-bit post-training quantization is more than 250x. This is an enormous decrease in the size of the model, with minuscule change in the accuracy of the model.

\section{Conclusions}
Performance of all of the methods have been summarized in Table \ref{tab:all_methods}.

\input{latex/tables/compress_all}

In general, we noticed that for almost all techniques, accuracy of the models improved after compression. This might have been due to our models being larger than the easy and small data sets we tested them on. Although, our experiments points towards excellent performance efficacy for the compressed models, the results might change when tested on much deeper networks and on large hard data sets. 

The other interesting conclusion we can draw from the knowledge distillation experiments is the advantage of introducing teacher assistant models when the gap between the student and the teacher is large. Although multi-step distillation helps in these cases, it will be interesting to see the performance for deeper models and for more than one steps in between teacher and student. Again, our conclusion is very rudimentary and these need to be explored further. 

Finally, from our experiments, we do notice the huge benefit of combining distillation with pruning and quantization. We can achieve highly compressed models without much change in performance. 

As part of future work, we would like to explore harder data sets with many more classes, test on deeper models, and experiment on combining other modern compression techniques. We tried to explore the very basic compression techniques in a simple manner, but many more sophisticated methodologies have been explored in the literature. Another part of potential future work could be to employ these algorithms on commonly used pre-existing networks, such as AlexNet, or even on much larger networks where network compression may actually be a problem, like DenseNet. Furthermore, we would like to experiment with even smaller student models in the future, to see just how compact a network can get while still performing well. It would also be interesting to work on an algorithm for recursively removing layers of a deep network in a manner analogous to backwards regression. A comprehensive and thorough analysis of all would be highly desirable. This project serves as a stepping stone in that direction.

\newpage
{\small
\bibliographystyle{ieee_fullname}
\bibliography{egbib}
}

\end{document}

%% file: latex/tables/global_prune.tex
\begin{table}[htb]
\centering
\begin{tabular}{|c|cc|cc|}

\hline
\multicolumn{1}{|l|}{\textbf{}}                       & \multicolumn{2}{c|}{\textbf{MNIST}} & \multicolumn{2}{c|}{\textbf{CIFAR-10}} \\ \hline
\textbf{Model}          & \multicolumn{1}{c|}{\textbf{Acc}} & \textbf{Size(MB)} & \multicolumn{1}{c|}{\textbf{Acc}} & \textbf{Size(MB)} \\ \hline
PD 50\%--80\%                           & \multicolumn{1}{c|}{+0.48}  & -68 & \multicolumn{1}{c|}{\textbf{+4.57}}   & \textbf{-68}   \\ \hline
PD 0\%--80\%                           & \multicolumn{1}{c|}{\textbf{+0.77}}  & \textbf{-68} & \multicolumn{1}{c|}{+3.27}   & -68   \\ \hline
PD 50\%--80\%                           & \multicolumn{1}{c|}{+0.62}  & -37 & \multicolumn{1}{c|}{+1.68}   & -37   \\ 
\hline
\hline
CS 20\%  & \multicolumn{1}{c|}{+0.54}  & -11.5 & \multicolumn{1}{c|}{+1.28}   & -11.4   \\ \hline
CS 50\%  & \multicolumn{1}{c|}{+0.65}  & -37 & \multicolumn{1}{c|}{+2.83}   & -37   \\ \hline
CS 80\%  & \multicolumn{1}{c|}{\textbf{+0.86}}  & \textbf{-68} & \multicolumn{1}{c|}{\textbf{+3.44}}   & \textbf{-68}   \\ \hline

\end{tabular}
\caption{Results of global unstructured pruning on the two baselines. All numbers show the percentage improvement over the original baseline models.}
\label{tab:glob_pruning}
\end{table}

%% file: latex/tables/local_prune.tex
\begin{table}[htb]
\centering
\begin{tabular}{|c|cc|cc|}

\hline
\multicolumn{1}{|l|}{\textbf{}}                       & \multicolumn{2}{c|}{\textbf{Dense Layers}} & \multicolumn{2}{c|}{\textbf{Conv Layers}} \\ \hline
\textbf{Model}          & \multicolumn{1}{c|}{\textbf{Acc}} & \textbf{Size(MB)} & \multicolumn{1}{c|}{\textbf{Acc}} & \textbf{Size(MB)} \\ \hline
PD 50\%--80\%   & \multicolumn{1}{c|}{\textbf{+3.8}}  & \textbf{-68} & \multicolumn{1}{c|}{+3.9}   & -68   \\ \hline
PD 0\%--80\% & \multicolumn{1}{c|}{\textbf{+3.8}}  & \textbf{-68} & \multicolumn{1}{c|}{\textbf{+4.2}}   & \textbf{-68}   \\ \hline
PD 50\%--80\%   & \multicolumn{1}{c|}{+2.6}  & -37 & \multicolumn{1}{c|}{+2.9}   & -37   \\ 
\hline
\hline
CS 20\%  & \multicolumn{1}{c|}{+1.2}  & -11.5 & \multicolumn{1}{c|}{+1.03}   & -11.4   \\ \hline
CS 50\%  & \multicolumn{1}{c|}{+2.7}  & -37 & \multicolumn{1}{c|}{+1.92}   & -37   \\ \hline
CS 80\%  & \multicolumn{1}{c|}{\textbf{+3.6}}  & \textbf{-68} & \multicolumn{1}{c|}{\textbf{+4.4}}   & \textbf{-68}   \\ \hline

\end{tabular}
\caption{Results of local unstructured pruning on the dense layers and convolutional layers separately for the CIFAR-10 baseline. All numbers show the percentage improvement over the original baseline models.}
\label{tab:local_pruning}
\end{table}

%% file: latex/tables/quantization.tex
\begin{table*}[htb]
\centering
\begin{tabular}{|c|cc|cc|}

\hline
\multicolumn{1}{|l|}{\textbf{}}                       & \multicolumn{2}{c|}{\textbf{MNIST}} & \multicolumn{2}{c|}{\textbf{CIFAR-10}} \\ \hline
\textbf{Model}          & \multicolumn{1}{c|}{\textbf{Accuracy}} & \textbf{Size (MB)} & \multicolumn{1}{c|}{\textbf{Accuracy}} & \textbf{Size (MB)} \\ \hline
Original Model & \multicolumn{1}{c|}{97.74}    & 5.923     & \multicolumn{1}{c|}{70.42}    & 5.22      \\ \hline
Quantization Aware Training                           & \multicolumn{1}{c|}{98.12}  & 1.494 & \multicolumn{1}{c|}{69.55}   & 1.332   \\ \hline
Model + 8-bit Post Training Quantization              & \multicolumn{1}{c|}{97.75}  & 1.494 & \multicolumn{1}{c|}{70.36}   & 1.326   \\ \hline
Model + 16-bit Post Training Quantization             & \multicolumn{1}{c|}{97.74}  & 2.964 & \multicolumn{1}{c|}{70.42}   & 2.614   \\ \hline
Distilled Model (Teacher \(\rightarrow\) Student)                   & \multicolumn{1}{c|}{97.61}  & 0.081 & \multicolumn{1}{c|}{70.62}   & 1.315   \\ \hline
Quantization Aware Teacher \(\rightarrow\) Distilled Student        & \multicolumn{1}{c|}{97.61}  & 0.081 & \multicolumn{1}{c|}{70.62}   & 1.315   \\ \hline
Teacher \(\rightarrow\) Quantization Aware Student                  & \multicolumn{1}{c|}{97.55}  & 0.024 & \multicolumn{1}{c|}{67.93}   & 0.345   \\ \hline
Teacher \(\rightarrow\) Student + 8-bit Post Training Quantization  & \multicolumn{1}{c|}{97.62}  & 0.023 & \multicolumn{1}{c|}{70.59}   & 0.341   \\ \hline
Teacher \(\rightarrow\) Student + 16-bit Post Training Quantization & \multicolumn{1}{c|}{97.61}  & 0.042 & \multicolumn{1}{c|}{70.62}   & 0.662   \\ \hline
\end{tabular}
\caption{Results of all Quantization Experiments}
\label{tab:quant_results}
\end{table*}

%% file: latex/tables/kd_prune.tex
\begin{table*}[htb]
\centering
\begin{tabular}{|c|cc|cc|}

\hline
\multicolumn{1}{|l|}{\textbf{}}                       & \multicolumn{2}{c|}{\textbf{MNIST}} & \multicolumn{2}{c|}{\textbf{CIFAR-10}} \\ \hline
\textbf{Model}          & \multicolumn{1}{c|}{\textbf{Accuracy}} & \textbf{Size (MB)} & \multicolumn{1}{c|}{\textbf{Accuracy}} & \textbf{Size (MB)} \\ \hline
Original (Teacher) & \multicolumn{1}{c|}{97.83}    & 5.76     & \multicolumn{1}{c|}{69.68}    & 5.08      \\ \hline
Teacher \(\rightarrow\) Pruning                           & \multicolumn{1}{c|}{98.29}  & 1.84 & \multicolumn{1}{c|}{72.87}   & 1.62   \\ \hline
Teacher \(\rightarrow\) Student             & \multicolumn{1}{c|}{97.75}  & 0.08 & \multicolumn{1}{c|}{70.34}   & 1.28   \\ \hline
Teacher \(\rightarrow\) Pruning \(\rightarrow\) Student      & \multicolumn{1}{c|}{97.75}  & 0.08 & \multicolumn{1}{c|}{71.40}   & 1.28   \\ \hline
Teacher \(\rightarrow\) Student \(\rightarrow\) Pruning   & \multicolumn{1}{c|}{96.94}  & 0.03 & \multicolumn{1}{c|}{72.16}   & 0.42   \\ \hline

\end{tabular}
\caption{Results of experiments combining knowledge distillation and global unstructured pruning on the two baselines.}
\label{tab:kd_pruning}
\end{table*}

%% file: latex/tables/compress_all.tex
\begin{table}[htb]
\centering
\begin{tabular}{|c|c|c|}

\hline
\multicolumn{1}{|c|}{\textbf{Model}}                       & \multicolumn{1}{c|}{\textbf{MNIST}} & \multicolumn{1}{|c|}{\textbf{CIFAR-10}} \\ \hline
Original (Teacher) & 16.84    & 13.72 \\
\hline
Knowledge Distillation & 1219.13    & 54.87 \\
\hline
KD with Teaching Assitant & 1220.13 & 54.66 \\
\hline
Global Pruning & 53.34 & 44.98 \\
\hline
Local Pruning (Dense) & 52.66 & 44.38 \\
\hline
Local Pruning (Conv) & 53.21 & 44.61 \\
\hline
Quant. Aware Training & 65.68 & 52.21 \\
\hline
KD + Pruning & \textbf{3258.33} & \textbf{171.81} \\
\hline
KD + 8-bit Post-train Quant. & \textbf{4244.35} & \textbf{207.01} \\
\hline

\end{tabular}
\caption{Performance efficacy (Accuracy/Size) of all the methods compared in this paper when tested on both the datasets.}
\label{tab:all_methods}
\end{table}